\documentclass[sigconf]{acmart}

\AtBeginDocument{%
  }

\copyrightyear{2022}
\acmYear{2022}
\setcopyright{acmcopyright}\acmConference[MM '22]{Proceedings of the 30th ACM International Conference on Multimedia}{October 10--14, 2022}{Lisboa, Portugal}
\acmBooktitle{Proceedings of the 30th ACM International Conference on Multimedia (MM '22), October 10--14, 2022, Lisboa, Portugal}
\acmPrice{15.00}
\acmDOI{10.1145/3503161.3548267}
\acmISBN{978-1-4503-9203-7/22/10}

\begin{document}

\title{Domain Generalization via Frequency-domain-based Feature Disentanglement and Interaction}

\author{Jingye Wang}
\orcid{0000-0003-3490-2363}
\affiliation{%
  \institution{Beijing University of Posts and Telecommunications}
  \country{China}
}
\email{wangjingye@bupt.edu.cn}

\author{Ruoyi Du}
\orcid{0000-0001-8372-5637}
\affiliation{%
  \institution{Beijing University of Posts and Telecommunications}
  \country{China}
}
\email{duruoyi@bupt.edu.cn}

\author{Dongliang Chang}
\orcid{0000-0002-4081-3001}
\affiliation{%
  \institution{Beijing University of Posts and Telecommunications}
  \country{China}
}
\email{changdongliang@bupt.edu.cn}

\author{Kongming Liang}
\orcid{0000-0002-4726-093X}
\authornote{Corresponding Author}
\affiliation{%
  \institution{Beijing University of Posts and Telecommunications}
  \country{China}
}
\email{liangkongming@bupt.edu.cn}

\author{Zhanyu Ma}
\orcid{0000-0003-2950-2488}
\affiliation{%
  \institution{Beijing University of Posts and Telecommunications}
  \country{China}
}
\email{mazhanyu@bupt.edu.cn}


\begin{abstract}
  Adaptation to out-of-distribution data is a meta-challenge for all statistical learning algorithms that strongly rely on the i.i.d. assumption. It leads to unavoidable labor costs and confidence crises in realistic applications. For that, domain generalization aims at mining domain-irrelevant knowledge from multiple source domains that can generalize to unseen target domains. In this paper, by leveraging the frequency domain of an image, we uniquely work with two key observations: (i) the high-frequency information of an image depicts object edge structure, which preserves high-level semantic information of the object is naturally consistent across different domains, and (ii) the low-frequency component retains object smooth structure, while this information is susceptible to domain shifts. Motivated by the above observations, we introduce (i) an encoder-decoder structure to disentangle high- and low-frequency feature of an image, (ii) an information interaction mechanism to ensure the helpful knowledge from both two parts can cooperate effectively, and (iii) a novel data augmentation technique that works on the frequency domain to encourage the robustness of frequency-wise feature disentangling. The proposed method obtains state-of-the-art performance on three widely used domain generalization benchmarks (Digit-DG, Office-Home, and PACS).
\end{abstract}


\begin{teaserfigure}
    \centering
    \includegraphics[width=17cm]{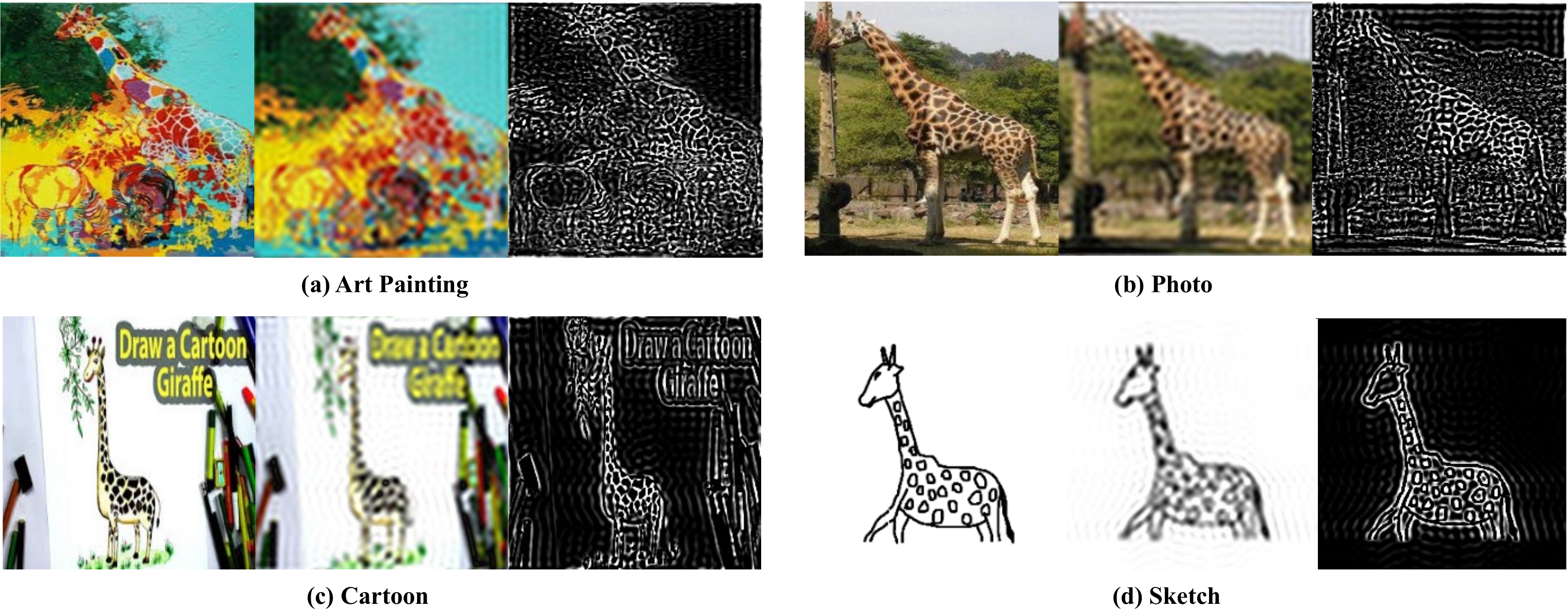}
    \caption{Examples of ``Giraffes'' in four domains from the PACS dataset. Each sub-figure is the original image, the low-pass filtered image, and the high-pass filtered image from left to right. We can see that the high-pass filtered versions retain the edge features of the object, which is less affected by the domain shift. And low-pass filtered versions retain image smooth contour information, which can assist with high-frequency information for classification.}
    \label{fig:1}
\end{teaserfigure}

\maketitle

\section{Introduction}
\label{sec1}
With years of vigorous development, deep learning methods have played important roles in various computer vision fields~\cite{du2021progressive,redmon2016you,xie2021gpca,simonyan2014very,chang2021your,chang2020devil,wei2019adversarial,wei2020lifelong}, which also have been heavily reliant on the i.i.d. assumption. However, the distribution of training data and test data often encounter significant shifts in practical applications (\emph{e.g.}, recognizing the same object in cartoons and art paintings), which may lead to sharp performance degradation of the deep neural network. Domain adaptation (DA)~\cite{peng2019moment,ganin2015unsupervised,ben2006analysis,wang2018visual,wang2020continuously} and domain generalization (DG) were proposed to relieve the domain distribution discrepancy. However, DA requires labeled or unlabeled data from the target domain to conduct the model adaptation, which makes it less practical in realistic scenarios. On the contrary, DG aims at training the model with multiple source domains so that it can be generalized to any target domains with unknown distributions. The existing DG methods can roughly divided into three trends~\cite{wang2021}: data manipulation~\cite{zhou2020learning,li2021,zhou2021,shankar2018generalizing,liu2018unified}, representation learning~\cite{wang2020learning,li2018,matsuura2020,liu2021domain,du2021learning}, and learning strategy~\cite{li2019,carlucci2019,dou2019domain,qiao2021uncertainty,sagawa2019distributionally}.

\begin{figure*}[t]
    \centering
    \includegraphics[width=\textwidth]{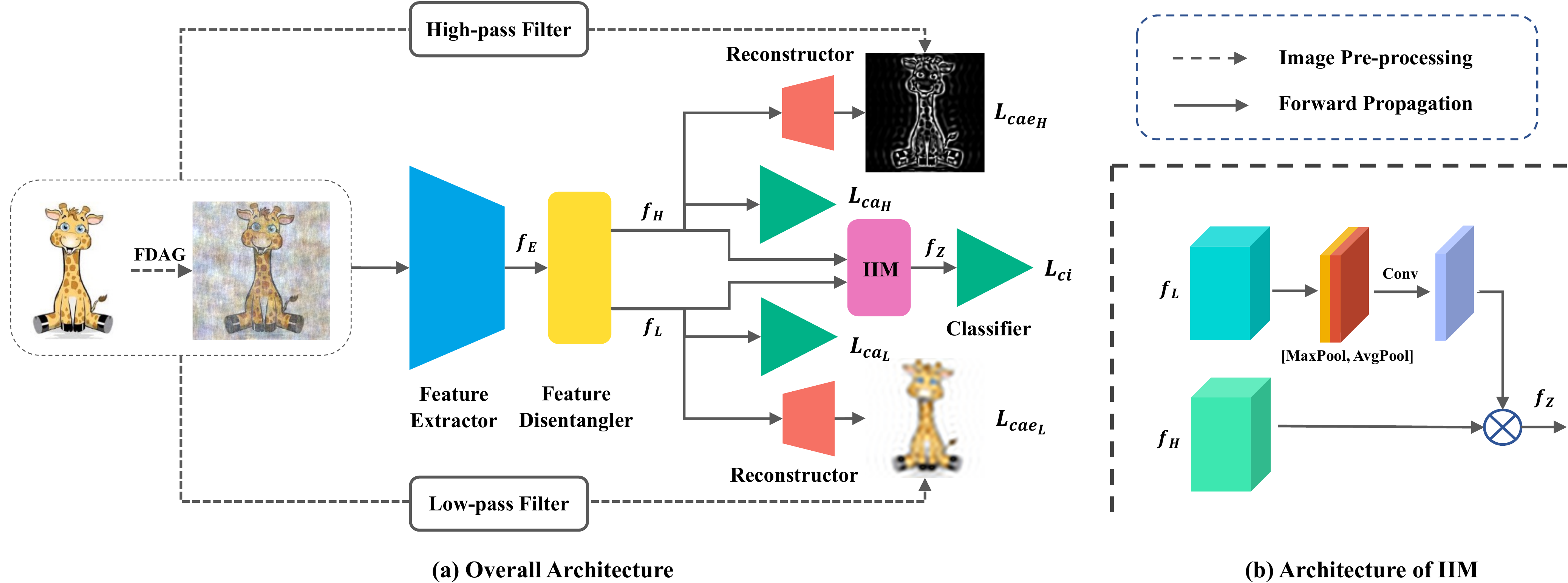}
    \caption{(a) The architecture of the FFDI. First, the original image is disturbed by FDAG, then the high- and low-frequency features are disentangled by CAE. After that, we interact between the disentangled frequency-specific features by IIM. Finally, we feed the fused feature to the classifier to obtain the prediction results. (b) The architecture of the IIM. We use low-frequency features to create a spatial mask and multiply it by high-frequency features to highlight the key information and suppress insignificant features.}
    \label{fig:2}
\end{figure*}

\begin{table}[t]
\centering
\caption{A-Distance of high- and low-frequency images.}
\label{tab:0}
\centering
\begin{tabular}{c|c}
\hline
Frequency & A-distance(Ave.)\\
\hline
L & $1.97$ \\
\hline
H & $1.87$ \\
\hline
\end{tabular}
\end{table}

Nevertheless, these previous arts mainly focus on how to learn with inter-domain interpolation or mine domain-irrelevant knowledge, but less or no efforts are paid for investigating the very question of which components of images carry the semantic information shared across domains. In this paper, instead of leaving everything to the model to learn, we tackle domain generalization problems with an empirical prior knowledge. In particular, we uniquely work with image frequency domain under two observations: (i) the high-frequency information of images depict object edge structure, which is naturally consistent across different domains~\cite{xu2021}, and (ii) the low-frequency component contains object smooth structure that retains more energy distributions of images but is much more domain-specific~\cite{yang2020fda,huang2021}. To further prove it, we calculate the A-distance~\cite{long2015learning} of high- and low-frequency images between every two domains on the PACS dataset~\cite{li2017} using ResNet18 features and report the average results in Tab.~$\ref{tab:0}$, which shows the high-frequency images are more similar between different domains than low-frequency images. To exploit the above observations, we perform the Fast Fourier Transform (FFT) on images of the same category from the PACS dataset to obtain their high-pass filtered versions and low-pass filtered versions as shown in Fig.~\ref{fig:1}. We can find that the high-pass filtered components from different domains are less affected by the domain shifts, and they are the domain-irrelevant semantics we hope to preserve. However, high-pass filtered images, which indicate where the gray value changes on a large scale~\cite{juneja2009}, lose most of the energies of the image and are more likely to be affected by noises caused by background edge structures (\emph{i.e.,} making it hard to distinguish the foreground and the background). On the contrary, although low-pass filtered images are much more domain-specific, they still retain most of the image energies, (\emph{i.e.,} object location information). Hence, we argue that low-frequency information can be complementary to high-frequency information and assist with high-frequency information while performing recognition. Taking Fig.~\ref{fig:1}(b) as an example, it is difficult to distinguish the giraffe in the woods only by high-pass filtered image, and low-pass filtered image can provide us with its outline.

To realize the above idea, we introduce a frequency-domain-based feature disentanglement and interaction (FFDI) framework, which consists of three modules: an encoder-decoder structure, the information interaction mechanism (IIM), and the frequency-domain-based data augmentation technique (FDAG), as shown in Fig.~\ref{fig:2}(a). First of all, we design a two branch encoder-decoder structure, which leverages the idea of feature disentanglement~\cite{peng2019} to obtain the high- and low-frequency features of an image. Second, we aim to effectively compose the semantic information delivered by high-frequency features and energy information represented by low-frequency features. For that, bilinear pooling~\cite{lin2015} tends to be a natural choice -- it is related to the dual-stream hypothesis of visual processing in the human brain~\cite{goodale1992separate}, which has two main pathways, or "streams": the ventral stream (or "what pathway") is involved with semantic information, and the dorsal stream (or "where pathway") involves spatial location information. However, bilinear pooling suffers from the problem of high dimensionality of the fused features, which leads to higher memory usage and time cost. Therefore, we introduce a simple yet effective feature interaction technique as shown in Fig.~\ref{fig:2}(b) with the similar insight to bilinear pooling.
Besides, we also try some other feature interaction methods: feature addition and feature concatenation. The experimental results in Tab.~$\ref{tab:5}$ show that our proposed method can be as effective as bilinear pooling in combining helpful information in the high- and low-frequency features and reduce memory consumption and computational complexity simultaneously. We can also find that even the simple feature interaction methods (\emph{e.g.}, addition, concatenation) can achieve good results, which proves the interaction between different frequency components of the images is effective. Finally, to further improve the robustness of the network to extract the high- and low-frequency features, we propose a frequency-domain-based data augmentation. The proposed data augmentation method is inspired by the recent works that aim to solve the domain shifts problem based on frequency domain~\cite{yang2020fda,xu2021}.

The contributions of our work are summarized as follow:
\begin{itemize}
\item We propose a new network structure named FFDI that disentangles high- and low-frequency features of the image and fuses two types of feature via the information interaction mechanism (IIM), which can exploit useful information from different frequency components to improve the representation power of the model.
\item To boost the dependability of feature disentanglement, we introduce a data augmentation method that executes noise interference in the frequency domain.
\item We demonstrate the efficacy of our proposed method on three standard domain generalization benchmarks (Digit-DG, Office-Home, and PACS) and obtain state-of-the-art performance. Moreover, extensive ablation studies are implemented to prove the validity of the method design.
\end{itemize}

\section{Related Work}
\textbf{Domain Generalization:} 
Over the years, tremendous progress has been made in the domain generalization field, which can be roughly divided into three trends~\cite{wang2021}. For data manipulation,~\citet{li2021} proposes a data augmentation method for embedding features during training, which can be combined with some existing technologies to obtain better generalization performance. 
For representation learning, ~\citet{wang2020learning} proposes EISNet that uses extrinsic relationship supervision and intrinsic self-supervision for images to implement domain generalization.
For learning strategy,~\cite{li2019} proposes an episodic learning strategy to obtain a robust baseline.

Our work is relevant to the frequency-domain-based DG methods. Inspired by some DA algorithms~\cite{yang2020fda,yang2020phase},~\citet{xu2021} argue that the phase information of images contains high-level semantic information and develops a Fourier-based data augmentation strategy by mixing the amplitude parts of two images aims to force the model to capture phase information.~\citet{huang2021} improves the generalization of the network by keeping the domain invariant frequency components and randomizing the domain variant frequency components in the segmentation field.~\citet{jeon2021feature} decomposes high- and low-frequency components from the embedding feature and then manipulates low-frequency features while preserving their high-frequency features, which can prevent semantic distortion. Different from previous algorithms, in this paper, we disentangle the high- and low-frequency features at the level of the image by using the encoder-decoder structure and then use an information interaction mechanism to interact between them to obtain more generalized features. Finally, we generalize the frequency-domain-based data augmentation method to improve the robustness of the network in extracting corresponding frequency features.\\

\noindent\textbf{Representation Disentanglement:}
Representation disentanglement is an active research topic in the field of computer vision. In face recognition, \citet{wang2020cross} proposes to disentangle PAD informative features and subject discriminative features by using a pair of encoders which is learned by generative models. In image translation, \citet{gonzalez2018} extracts content space and style space from the latent space of the image by using a pair of Generative
Adversarial Networks (GANs). Aims to improve the generalization capabilities. 
\citet{peng2019} designs a deep adversarial disentangled auto-encoder (DADA) to disentangle domain-specific features from class identity. In this paper, we disentangle the high- and low-frequency of the image in the latent space by using an encoder-decoder structure.

\section{Methodology}

We define the domain generalization task as follows: given $K$ source domains $ D_S = \{D_1,...,D_K\}$ as the training set, where the $i$-th domain $D_i$ have $N_i$ image-label sample pairs $\{(x_{j}^{i}, y_{j}^{i})\}_{j=1}^{N_i}$. The target is to learn a model from multiple source domains that can be generalized to the target domains $D_T$ with unknown distributions.

In this paper, to relieve the problem of network performance degradation caused by domain shifts, we propose a frequency-domain-based feature disentanglement and interaction (FFDI) framework. The overall network structure of our method is illustrated in Fig.~\ref{fig:2}(a). The feature extractor $E(\cdot)$ maps the input image to the embedding feature $f_E$. The disentangler $D(\cdot)$ is composed of two parallel convolutional layers and is responsible for disentangling $f_E$ into high-frequency features $f_H$ and low-frequency features $f_L$. The image reconstructor $R(\cdot)\in \{R_H(\cdot), R_L(\cdot)\}$ aims to recover high-pass (or low-pass) filtered images from $f_H$ (or $f_L$). Hence $D(\cdot)$ and $R(\cdot)$ can be 
viewed as the encoder-decoder structure in Convolution Auto-Encoders (CAE). To effectively utilize helpful information in high- and low-frequency features, we design a  information interaction mechanism $IMM(\cdot)$ for the fusion of $f_H$ and $f_L$. After that, we use three classifiers $C_{A_H}(\cdot)$, $C_{A_L}(\cdot)$, $C_I(\cdot)$ corresponding to high-frequency features, low-frequency features, and fused features respectively to predict the target category. Note that $C_{A_H}(\cdot)$ and $C_{A_L}(\cdot)$ only work as auxiliary classifiers to enhance the discriminative information in $f_H$ and $f_L$ and will be discard during inference. Furthermore, we propose a frequency-domain-based data augmentation method named FDAG, which can enhance the robustness of the feature disentanglement. Next, we will describe each module of FFDI in detail.

\subsection{Feature Disentanglement}
\label{sec3.1}
We use CAE to extract high- and low-frequency features of the image. First of all, we transform each channel of the RGB image $I$ to the frequency domain space by using the Fourier transformation, which is formulated as:
\begin{align}
\label{Eq.1}
    F(u,v) = \sum_{a=0}^{A-1} \sum_{b=0}^{B-1} x(a,b) e^{-j2\pi(au/A + bv/B)},
\end{align}
and for simplified we donate Fourier transformation as $F(\cdot)$ and use the symbol $F^{-}(\cdot)$ to represent the inverse Fourier transformation. Further, we denote with $M$ a mask, whose value is zero except for the center region:

\begin{equation}
\label{eq6}
M={\left\{
\begin{aligned}
1 & , (u,v) \in [c_x-r:c_x+r, c_y-r:c_y+r] \\
0 & , others
\end{aligned}
\right.},
\end{equation}
where $(c_x, c_y)$ is the center of the image and $r$ indicates the frequency threshold that distinguishes between high- and low-frequencies of the original image. Then, the low-pass filtered image ($LFI$) and high-pass filtered image ($HFI$) can be obtained as follows:
\begin {align}
LFI &= F^{-}(F(I) \circ M),\\ 
HFI &= I - LFI,
\end {align}
where $\circ$ denotes the Hadamard product of the matrix.
And the obtained $LFI$ and $HFI$ are used as the CAE's optimization targets. For training the disentangler $D(\cdot)$ and reconstructor $R(\cdot)$ to correctly reconstruct the $HFI$ and $LFI$, the reconstructed loss of the CAE over multiple source domains is defined as
\begin {align}
\mathcal{L}_{cae} = \frac{1}{K} {\sum_{i=1}^{K}} \frac{1}{N_i} \sum_{j=1}^{N_i} {\Vert {X_{f_{j}}^i - \hat{X}_{f_{j}}^i}}\Vert_{2}^{2},
\end {align}
where $X_{f} \in \{LFI, HFI\}$, $\hat{X}_{f}$ is the output of $R(\cdot)$.

By using $HFI$ and $LFI$ as target labels for high-pass filtered image and low-pass filtered image reconstruction respectively, we can make the embedded features $f_H$ or $f_L$ more biased towards high-frequency features or low-frequency features of the image.

Meanwhile, we employ the $f_H$ and $f_L$ as input to train the auxiliary classifier ($C_{A_H}$, $C_{A_L}$) to correctly predict the sample class, which allows frequency-specific features to obtain corresponding semantic information.
This can be achieved by minimizing the standard cross-entropy loss:%
\begin{align}
    \mathcal{L}_{ca} = - \frac{1}{K} \sum_{i=1}^K \frac{1}{N_i} \sum_{j=1}^{N_i} {y_j^i} \log(C_A(AvgPool(f_{F_j}^i))),
\end{align}%
where $f_F \in \{f_H, f_L\}$, $C_A \in \{C_{A_H}, C_{A_L}\}$.

\subsection{Information Interaction Mechanism}
As analyzed in Sec.~\ref{sec1}, to make full use of the helpful information of both $f_H$ and $f_L$, we establish a practical information interaction mechanism. Inspired by \cite{lin2015,woo2018cbam}, we use the $f_L$ extracted in Sec.~\ref{sec3.1} to generate the corresponding spatial masks and multiply them with the $f_H$ to encode where to emphasize or suppress. 

To be specific, for features $f_L\in R^ {C\times H\times W}$ and $f_H \in R^ {C\times H\times W}$ extracted from the same sample, where $C, H, W$ respectively denote the number of channels, height, and width of the feature map, first, we use average-pooling and max-pooling for $f_L$ along the channel axis to obtain $f_{L_{avg}}\in R^ {1\times H\times W}$  and $f_{L_{max}}\in R^ {1\times H\times W}$, and then concatenate them to generate the feature $f_{spatial}\in R^ {2\times H\times W}$ that can effectively highlight spatial information. After that, the feature $f_{spatial}$ is fed into a standard convolutional layer to obtain 2D spatial mask $f_{mask}$. Formula is as follows:
\begin{align}
   f_{mask}=\sigma(Conv([AvgPool(f_L),MaxPool(f_L)])),
\end{align}
where $\sigma$ denotes sigmoid function and $Conv$ denotes a standard convolutional layer. Then we multiply the spatial mask $f_{mask}$ by $f_H$ to obtain the fused feature $f_Z$. 

After that, we use $f_Z$ as the input of classifier $C_I$ consisting of a fully connected layer to correctly predict the class of the each image, which is supervised by the cross-entropy loss:
\begin{align}
    \mathcal{L}_{ci} = -\frac{1}{K} \sum_{i=1}^K \frac{1}{N_i} \sum_{j=1}^{N_i} {y_j^i} \log(C_I(AvgPool(f_{Z_j}^{i}))).
\end{align}%

Intuitively, by jointly training $f_H$ and $f_L$, our method can trade off the ability of data-based feature representation, We hope to use the interaction between $f_H$ and $f_L$ for mutual constraints, 
which may allow the network to learn object edge features in high-frequency features while also noticing helpful information in low-frequency features. It is worth noting that our method is simple but has a significant effect on the generalization of the network, which proves the validity of the idea of interaction between high- and low-frequency features.

\begin{figure}[t]
    \centering
    \includegraphics[width=7cm]{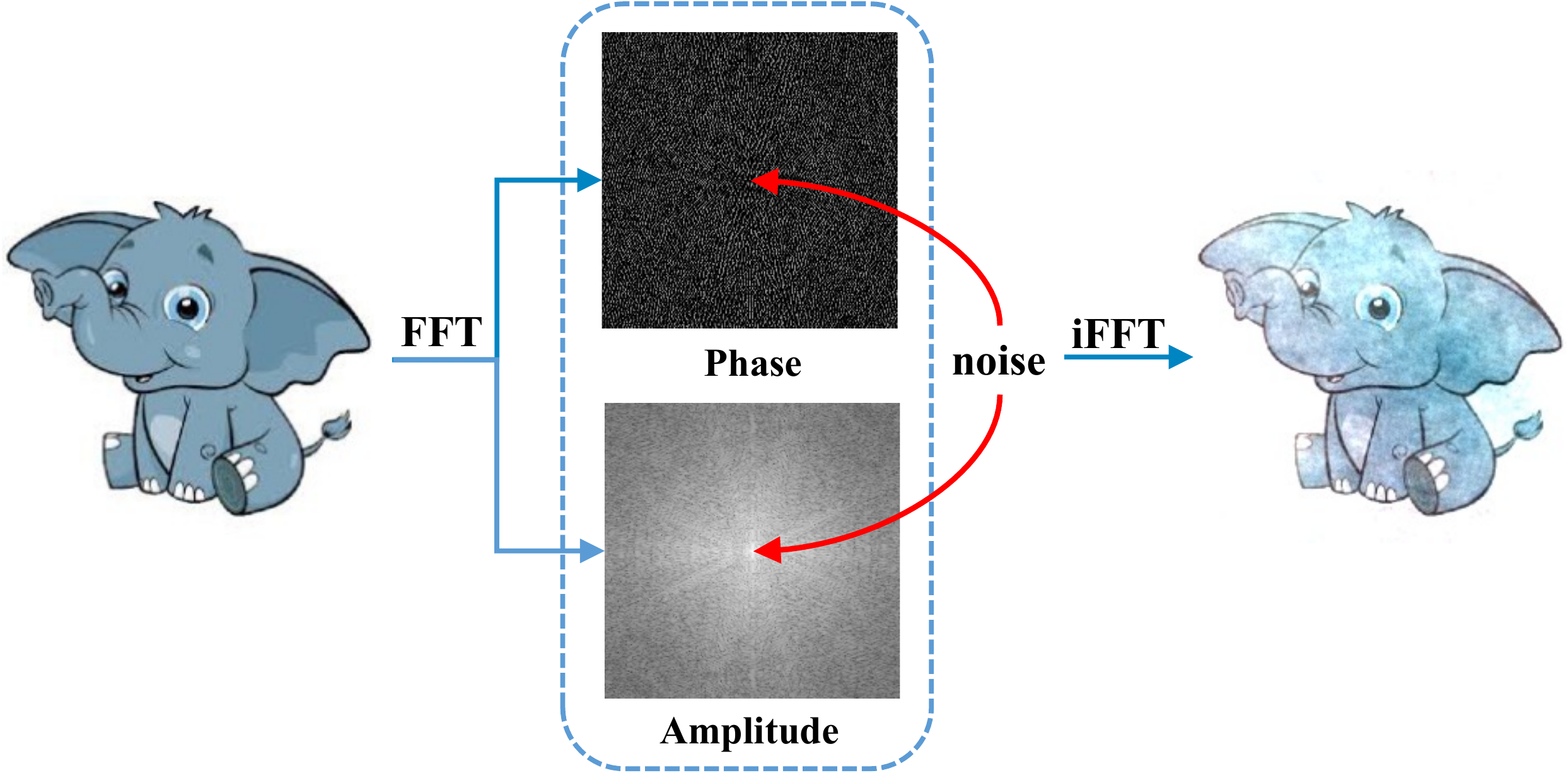}
    \caption{Frequency-domain-based data augmentation. Multiplicative and additive noises are both applied to the phase and amplitude of the image's frequency domain.}
    \label{fig:3}
\end{figure}

\subsection{Frequency-domain-based Data Augmentation}
The above method can make a large improvement in the generalization ability of the network, however, we cannot guarantee that it can also extract the high- and low-frequency features of the image well on the unseen domain, which will directly affect the robustness of the model. To alleviate this problem, intuitively, we propose a simple but effective data augmentation technique that works on the frequency domain.

First, we obtain the frequency domain representation of the image using Eq.~\ref{Eq.1}. And then we convert it to polar coordinate form:
\begin{align}
    F(u,v) = |F(u,v)| e^{-j\phi(u,v)},
\end{align}
then we can obtain the mathematical expressions for its amplitude and phase:
\begin{align}
    A(u,v) &= |F(u,v)|, \\
    P(u,v) &= \phi(u,v).
\end{align}%

In this paper, we apply frequency domain disturbance to enrich the diversity of sample distribution. Many previous work~\cite{xu2021,yang2020fda} argue the phase of the image has high-level semantic information, so they are both working on the amplitude such as exchanging the amplitude information between different images without changing the phase. We generalize this method by simply applying random noise to amplitude. In addition, we find that a slight disturbance to the phase can further improve the generalization ability of the network, and the experimental results are listed in Tab.~$\ref{tab:6}$. Specifically, inspired by \cite{li2021}, we utilize multiplicative and additive noise in the phase and amplitude. The computation formula is as follows:
\begin{align}
    \hat{A_g} = \alpha \circ A_g + \beta,
\end{align}%
where $A_g \in \{P, A\}$, 
$\alpha \in R^{C\times H\times W}$ and $\beta \in R^{C\times H\times W}$ are multiplicative noise and additive noise, respectively. For example, each element $\alpha$ is sampled from Uniform distributions $U(a,b)$ and $\beta$ is sampled from Normal distributions $N(\mu,\sigma ^2)$. We then feed $\hat{A_g}$ into the inverse Fourier transform to obtain the augmented image. The frequency-domain-based data augmentation technique (FDAG) is shown in Fig.~\ref{fig:3}.

\subsection{Algorithm Flow}
The above three components together form the FFDI framework. In the training stage, we first augment the data using FDAG, then feed the data to the network
and use the overall loss $\mathcal{L}_{all}$ to train our model as follow:
\begin{align}
\label{Eq.15}
    \mathcal{L}_{all} = \mathcal{L}_{ci} + \lambda(\mathcal{L}_{ca_{L}} + \mathcal{L}_{ca_{H}} +  \mathcal{L}_{cae_L} + \mathcal{L}_{cae_H}).
\end{align}

After training, only $E(\cdot)$, $D(\cdot)$, $IIM(\cdot)$, and $C_I(\cdot)$ will be deployed for inference.

\section{Experiments}

\subsection{Datesets and Settings}
\textbf{Datasets:} To validate the performance of the proposed method FFDI, we conduct extensive experiments on the following datasets: \textbf{PACS}~\cite{li2017} that consists of four domains with different data distribution (\textit{Photo}, \textit{Art painting}, \textit{Cartoon}, and \textit{Sketch}) where each domain contains $7$ categories, and the whole dataset has $9,991$ images. \textbf{Digits-DG}~\cite{zhou2020learning} contains four datasets (\textit{MNITS}, \textit{MNIST-M}, \textit{SVHN}, and \textit{SYN}) in the field of digital handwriting recognition with diverse image backgrounds and styles. Following the experimental protocol of \cite{zhou2020learning}, each domain has $600$ images per category, and the ratio of the training and test sets in the source domain is $4:1$. \textbf{Office-Home}~\cite{venkateswara2017} is composed of four domains (\textit{Art}, \textit{Clipart}, \textit{Product}, and \textit{Real world}) with a total of $15,500$ images and $65$ categories. This dataset contains images with various styles and viewpoints. Following the setting of \cite{zhou2020learning}, we split the dataset according to $\text{training}: \text{test} = 9:1$. 


\noindent\textbf{Evaluation Protocol:} To make a fair comparison, we follow \cite{zhou2020learning,venkateswara2017} to apply the leave-one-domain-out protocol, which means selecting one domain as the unseen domain and the rest of the domains are considered source domains to train our network. In this paper, we report the top-1 classification accuracy on the target domains, which is averaged over three runs. We use the vanilla convolutional neural network trained on the simple aggregation of all source domains as our baseline named DeepAll. We also used the standard augmentation protocol in~\cite{carlucci2019}, which includes random horizontal flipping and color jittering.

\begin{table}[t]
\caption{Leave-one-domain-out results on PACS datasets with ResNet18 and ResNet50. The best results are bolded and the sub-optimal results are underlined.}
\label{tab:1}
\begin{tabular}{c|c|c|c|c|c}
\hline
Target  &Art &Cartoon &Photo &Sketch & Ave.\\
\hline
\multicolumn{6}{c}{ResNet18} \\
\hline
DeepAll & $79.6$ & $76.0$ & \underline{$96.4$} & $66.9$ & $79.7$  \\
MetaReg~\cite{balaji2018} & $83.7$ & $77.2$ & $95.5$ & $70.3$ & $81.7$ \\
Epi-FCR~\cite{li2019} & $82.1$ & $77.0$ & $93.9$ & $73.0$ & $81.5$\\
JiGen~\cite{carlucci2019} & $79.4$ & $75.3$ & $96.0$ & $71.4$ & $80.5$\\
CrossGrad~\cite{shankar2018} & $79.8$ & $76.8$ & $96.0$ & $70.2$ & $80.7$ \\
DDAIG~\cite{zhou2020deep} & $84.2$ & $78.1$ & $95.3$ & $74.7$ & $83.1$ \\
CSD~\cite{piratla2020} & $78.9$ & $75.8$ & $94.1$ & $76.7$ & $81.4$ \\
L2A-OT~\cite{zhou2020learning} & $83.3$ & $78.2$ & $96.2$ & $73.6$ & $82.8$  \\
MixStyle~\cite{zhou2021} & $84.1$ & $78.8$ & $96.1$ & $75.9$ & $83.7$ \\
RSC~\cite{huang2020} & $83.4$ & $80.3$ & $96.0$ & $80.9$ & $85.2$  \\
EISNet~\cite{wang2020learning} & $81.9$ & $76.4$ & $96.0$ & $74.3$ & $82.2$ \\
MDGH~\cite{mahajan2021} & $82.8$ & \textbf{81.6} & \textbf{96.7} & $81.1$ & $85.5$ \\
FSDCL~\cite{jeon2021feature} & \textbf{85.3} & $81.3$ & $95.6$ & \underline{$81.2$} & \underline{$85.9$} \\

\hline
FFDI(ours) & \underline{$85.2$} & \underline{81.5} & $95.8$ & \textbf{82.8}  & \textbf{86.3}\\
\hline
\multicolumn{6}{c}{ResNet50} \\
\hline
DeepAll & $86.3$ & $77.6$ & \underline{$98.2$} & $70.1$ & $83.0$ \\
MetaReg~\cite{balaji2018} & $87.2$ & $79.2$ & $97.6$ & $70.3$ & $83.6$ \\
CrossGrad~\cite{shankar2018} & $87.5$ & $80.7$ & $97.8$ & $73.9$ & $85.7$ \\
DDAIG~\cite{zhou2020deep} & $85.4$ & $78.5$ & $95.7$ & $80.0$ & $84.9$ \\
MixStyle~\cite{zhou2021} & $87.4$ & $83.3$ & $98.0$ & $78.5$ & $86.8$ \\
EISNet~\cite{wang2020learning} & $86.6$ & $81.5$ & $97.1$ & $78.1$ &$85.8$ \\
RSC~\cite{huang2020} & $87.9$ & $82.2$ & $97.9$ & \underline{$83.4$} & $87.8$\\
MDGH~\cite{mahajan2021} &$86.7$ &$82.3$ &\textbf{98.4}  &$82.7$ &$87.5$ \\
FSDCL~\cite{jeon2021feature} & \underline{$88.5$} & \underline{$83.8$} & $96.6$ & $82.2$ & \underline{$88.0$} \\
\hline
FFDI(ours) &\textbf{89.3} &\textbf{84.7} &$97.1$ &\textbf{83.9} &\textbf{88.8} \\
\hline
\end{tabular}
\end{table}

\subsection{Evaluation on PACS}
\textbf{Implementation Details:} We use the ResNet model pretrained with the ImageNet as our backbone of all experiments on this dataset. The network is trained with Stochastic Gradient Descent (SGD) optimizer. The batch size is $16$ per domain and $32$ per domain for ResNet$18$ and ResNet$50$ respectively. We set the weight decay as 1e-4 for $6000$ iterations. The $C_I$'s initial learning rate is $0.01$, and others are $0.001$, which decayed by $0.1$ at every $1000$ iterations. We set $\lambda=1$, $r=25$ and resize image to $224 \times 224$.

\noindent\textbf{Results:} We summarize the experiment results on PACS dataset in Tab.~$\ref{tab:1}$. Our FFDI achieves the best average performance among all the compared methods including the recent methods MDGH, FSDCL, and ATSRL. Sketch has a distinct difference in data-style from the remaining three domains (Art, Cartoon, and Photo), our performance still is almost $1.6\%$ higher than the second-best method FSDCL and also achieved satisfactory results on the remaining three domains. All the above results prove that our method can significantly improve the generalization ability of the network.

\begin{table}[t]
\centering
\caption{Leave-one-domain-out results on Digits-DG datasets.}
\label{tab:2}
\centering
\resizebox{0.47\textwidth}{!}{
\begin{tabular}{c|c|c|c|c|c}
\hline
Target &MNIST &MNIST-M &SVHN &SYN &Ave.\\
\hline

DeepAll & $96.6$ & $59.5$ & $61.8$ & $78.4$ & $74.1$    \\
CCSA~\cite{motiian2017} & $95.2$ & $58.2$ & $65.5$ & $79.1$ & $74.5$  \\
MMD-AAE~\cite{li2018} & $96.5$ & $58.4$ & $65.0$ & $78.4$ & $74.6$\\
JiGen~\cite{carlucci2019} & $96.5$ & $61.4$ & $63.7$ & $74.0$ & $73.9$\\
CrossGrad~\cite{shankar2018} & $96.7$ & $61.1$ & $65.3$ & $80.2$ & $75.8$ \\
DDAIG~\cite{zhou2020deep} & $96.6$ & \underline{$64.1$} & $68.6$ & $81.0$ & $77.6$ \\
L2A-OT~\cite{zhou2020learning} & \underline{$96.7$} & $63.9$ & \underline{$68.6$} & \underline{$83.2$} & \underline{$78.1$}  \\
MixStyle~\cite{zhou2021} & $96.5$ & $63.5$ & $64.7$ & $81.2$ & $76.5$ \\
\hline
FFDI(ours) & \textbf{97.7} & \textbf{69.4} & \textbf{72.1} & \textbf{84.5} & \textbf{80.9}  \\

\hline
\end{tabular}}
\end{table}

\subsection{Evaluation on Digit-DG}
\textbf{Implementation Details:} Following the experimental configuration in previous work~\cite{zhou2020learning}, we use the same backbone network, which is constructed by four $64$-kernels $3\times3$ convolution layers - ReLU - $2\times2$ max-pooling modules and a 
fully connected layer as the classifier that takes the flattened vector as input. The network is trained using Stochastic Gradient Descent (SGD) optimizer. The batch size is $42$ images per domain, and weight decay is $5$e-4 for $6000$ iterations. During training, our classifier's initial learning rate is set to $0.05$, and remain modules' learning rate is $0.01$, which decays by $0.1$ at every $1000$ iteration.We set $\lambda=1, r=3$ and resize image to $32 \times 32$.

\noindent\textbf{Results:} The results are reported in Tab.~$\ref{tab:2}$. We can see that our approach achieves the best average performance. On the two most difficult domains, MINIST-M and SVHN, The performance of our FFDI is $9.9\%$ and $10.3\%$ higher than the normal model, respectively. And on the remaining two domains, MNIST and SYN, our method also obtains competitive results. This again demonstrates the effectiveness of our proposed approach.

\subsection{Evaluation on Office-Home}
\textbf{Implementation Details:} We use the ResNet18 pretrained by ImageNet as our backbone of all experiments on this dataset. The network is trained with Stochastic Gradient Descent (SGD) optimizer. The batch size is $32$ per domain, and weight decay is 1e-4 for $6000$ iterations. The $C_I$'s initial learning rate is $0.01$, and others are $0.001$, which decays by $0.1$ at every 
$2000$ iterations. We set $\lambda=1/3$, $r=25$ and resize image to $224 \times 224$.

\noindent\textbf{Results:} The results are reported in Tab.~$\ref{tab:3}$. Our FFDI surpasses all the comparison methods in the table. We can see that DeepAll performs well on the Office-Home dataset, which is because the image number per category is small in this dataset and the data style is similar to its pretrained dataset ImageNet. Our method is further improved based on DeepAll and achieves the best average performance.

\begin{table}[t]
\caption{Leave-one-domain-out results on Office-Home datasets with ResNet18.}
\label{tab:3}
\resizebox{0.47\textwidth}{!}{
\begin{tabular}{c|c|c|c|c|c}
\hline
Target  &Art &Clipart &Product &Real Word &Ave.\\
\hline
DeepAll & $60.3$  & $50.4$  & $73.2$  & $75.3$ & $64.8$   \\
CCSA~\cite{motiian2017}  & $59.9$ & $49.9$ & $74.1$ & $75.7$ & $64.9$  \\
MMD-AAE~\cite{li2018} & $56.5$ & $47.3$ & $72.1$ & $74.8$ & $62.7$\\
JiGen~\cite{carlucci2019} & $53.0$ & $47.5$ & $71.5$ & $72.8$ & $61.2$\\
CrossGrad~\cite{shankar2018} & $58.4$ & $49.4$ & $73.9$ & $75.8$ & $64.4$ \\
DDAIG~\cite{zhou2020deep} & $59.2$ & $52.3$ & \underline{$74.6$} & $76.0$ & $65.5$ \\
RSC~\cite{huang2020} & $58.4$ & $47.9$ & $71.6$ & $74.5$ & $63.1$  \\
L2A-OT~\cite{zhou2020learning} & \underline{$60.6$} & $50.1$ & \textbf{74.8} & \textbf{77.0} & $65.6$  \\
MixStyle~\cite{zhou2021} & $58.7$ & $53.4$ & $74.2$ & $75.9$ & $65.5$\\
FSDCL~\cite{jeon2021feature} & $60.2$ & \underline{$53.5$} & $74.4$ & \underline{$76.7$} & \underline{$66.2$} \\
\hline
FFDI(ours) & \textbf{61.7} & \textbf{53.8} & $74.4$ & $76.2$ & \textbf{66.5}  \\
\hline
\end{tabular}}
\end{table}


\begin{table}[t]
\footnotesize
\centering
\caption{Ablation study on each component of FFDI on PACS with ResNet18.}
\label{tab:4}
\resizebox{0.47\textwidth}{!}{
\begin{tabular}{cccc|c|c|c|c|c}
\hline
Backbone & L & H & FDAG  &Art &Cartoon &Photo &Sketch &Ave.\\
\hline
\checkmark  & - & -  & -  & $79.6$ & $76.0$ & $96.4$ & $66.9$ & $79.7$\\
\checkmark  &\checkmark & -  & - & $79.1$ & $78.0$ & $91.1$ & $71.9$ & $80.0$\\
\checkmark  & - &\checkmark  & -  & $81.1$ & $77.2$ & $93.6$ & $73.0$ & $81.2$\\
\checkmark  &- &- &\checkmark  & $84.2$ & $76.1$ & \textbf{96.9} & $75.5$ & $83.2$\\
\checkmark  &\checkmark &\checkmark & -  & $82.6$ & $79.6$ & $95.3$ & $74.9$ & $83.1$\\
\hline
\checkmark  &\checkmark &\checkmark  &\checkmark     & \textbf{85.2} & \textbf{81.5} & $95.8$ & \textbf{82.8}  & \textbf{86.3}\\
\hline
\end{tabular}}
\end{table}

\subsection{Further Analysis}
\textbf{Impact of Different Components:} We describe an ablation study to investigate the effects of different components of FFDI using PACS and ResNet$18$. Some of the versions used in this experiment are as follows: ``H'' uses only high-frequency features for classification, which is optimized by loss functions $\mathcal{L}_{ca_H}$ and $\mathcal{L}_{cae_H}$.  ``L'' uses only low-frequency features for classification, which is optimized by loss functions $\mathcal{L}_{ca_L}$ and $\mathcal{L}_{cae_L}$. When we conduct interaction between ``H'' and ``L'' by using information interaction mechanism , our network is optimized by using Eq.~$\ref{Eq.15}$. ``FDAG'' indicates that Backbone is trained with our proposed frequency-domain-based data augmentation method FDAG. 

The results are reported in Tab.~$\ref{tab:4}$. When we use ``H'' or ``L'' alone, the result shows that ``L'' and baseline have similar performance, and ``H'''s performance is better than ``L'' and baseline, which proves that high-frequency features can improve the generalization of the network under the encoder-decoder structure. However, the performance of ``H'' and ``L'' in the photo domain is not as good as that of baseline, which is expected. As we analyzed in Sec.~\ref{sec1}, the photo domain scenario is the most complex in the PACS dataset, and it is impossible to obtain satisfactory results using ``H'' or ``L'' alone that both lose some information. Then we introduced the information interaction mechanism that interacts between high- and low-frequency features, which improves the average accuracy by $1.9\%$ on the ``H'' and the performance on the photo domain has been improved from $93.6\%$ to $95.3\%$. This indicates that fused features can effectively enhance the representational ability of the network. ``FDAG'' can increase the performance of baseline by $3.5\%$, which shows that FDAG can significantly enhance the robustness of the network. When we combine FDAG with our proposed information interaction method, the performance of our framework has a huge improvement (from $83.1\%$ to $86.3\%$), which validates FDAG plays an important role in the FFDI framework.

\begin{table}[t]
\centering
\caption{Comparision of different feature interaction methods on PACS with ResNet18.}
\label{tab:5}
\centering
\begin{tabular}{c|c|c|c|c|c}
\hline
Methods &Art &Cartoon &Photo &Sketch & Ave.\\
\hline
addition & $85.2$ & $81.0$ & $95.2$ & $81.1$ & $85.6$    \\
concatenation & $84.8$ & $80.1$ & \textbf{96.0} & $82.1$ & $85.7$  \\
bilinear pooling & \textbf{85.5} & $80.6$ & $95.5$ & $82.7$ & $86.1$\\
\hline
IIM & $85.2$ & \textbf{81.5} & $95.8$ & \textbf{82.8} & \textbf{86.3}  \\
\hline
\end{tabular}
\end{table}

\begin{table}[t]
\centering
\caption{Ablation study on each components of FDAG on PACS with ResNet18.}
\label{tab:6}
\centering

\begin{tabular}{c|c|c|c|c|c}
\hline
Methods &Art &Cartoon &Photo &Sketch & Ave.\\
\hline
\multicolumn{6}{c}{DeepAll with} \\
\hline

none & $79.6$ & $76.0$ & $96.4$ & $66.9$ & $79.7$
\\
phase & $82.7$ & \textbf{76.5} & $96.7$ & $70.8$ & $81.7$    \\
amplitude & $83.3$ & $76.4$ & $96.6$ & $74.3$ & $82.6$  \\

amplitude+phase & \textbf{84.2} & $76.1$ & \textbf{96.9} & \textbf{75.5} & \textbf{83.2}\\
\hline
\multicolumn{6}{c}{FFDI with} \\
\hline
none & $82.6$ & $79.6$ & $95.3$ & $74.9$ & $83.1$
\\
phase & $83.2$ & $81.1$ & $94.6$ & $79.4$ & $84.6$    \\
amplitude & $83.9$ & $80.2$ & $95.7$ & $82.1$ & $85.5$  \\
amplitude+phase & \textbf{85.2} & \textbf{81.5} & \textbf{95.8} & \textbf{82.8} & \textbf{86.3}\\
\hline
\end{tabular}
\end{table}

\noindent\textbf{Impact of Different Information Interaction Methods:} In order to verify the validity of our proposed idea based on high- and low-frequency feature interaction, we compare several common feature interaction methods: ``addition'', ``concatenation'', and ``bilinear pooling''. For ``addition'' and ``concatenation'' methods, we add and concatenate the extracted $f_H$ and $f_L$ respectively to obtain the fused features $f_Z$. ``bilinear pooling''~\cite{lin2015} is a classical feature interaction method in the field of fine-grained classification, and the insight in this paper has some similarities with it, so we try to apply this method to our framework. The results for different interaction methods are reported in Tab.~$\ref{tab:5}$, respectively. We find that the spatial mask-based information interaction mechanism proposed in this paper achieves the best performance, which is $0.2\%$ higher than bilinear pooling, but the computational complexity and memory are much smaller than ``bilinear pooling''. It is worth noting that although our method achieves the best performance, ``addition'' and ``concatenation'', two common feature interaction methods, also achieve good performance ($85.6\%$ and $85.7\%$ respectively), which illustrates the effectiveness of the idea of interaction between
high- and low-frequency features of the image.

\begin{table}[t]
\centering
\caption{Comparision of different data augmentation methods on PACS with ResNet18.}
\label{tab:7}
\centering
\resizebox{0.47\textwidth}{!}{
\begin{tabular}{c|c|c|c|c|c}
\hline
Methods &Art &Cartoon &Photo &Sketch & Ave.\\
\hline
\multicolumn{6}{c}{DeepAll with} \\
\hline
standard & $81.2$ & \textbf{77.6} & $96.6$ & $69.6$ & $81.3$ \\
t.-d. noise & $81.0$ & $76.7$ & $96.1$ & $69.3$ & $80.8$\\
AM & $83.7$ & $76.4$ & $95.8$ & \textbf{76.2} & $83.0$ \\
FDAG & \textbf{84.2} & $76.1$ & \textbf{96.9} & $75.5$ & \textbf{83.2} \\
\hline
\multicolumn{6}{c}{FFDI with} \\
\hline
standard & $82.3$ & $80.2$ & $95.6$ & $79.8$ & $84.5$    \\
t.-d. noise & $82.4$ & $80.7$ & $94.7$ & $78.6$ & $84.1$\\
AM & $85.1$ & $80.4$ & $95.2$ & $80.7$ & $85.4$ \\
FDAG & \textbf{85.2} & \textbf{81.5} & \textbf{95.8} & \textbf{82.8} & \textbf{86.3}  \\
\hline
\end{tabular}}
\end{table}

\begin{figure*}[t]
    \centering
    \includegraphics[width=11cm]{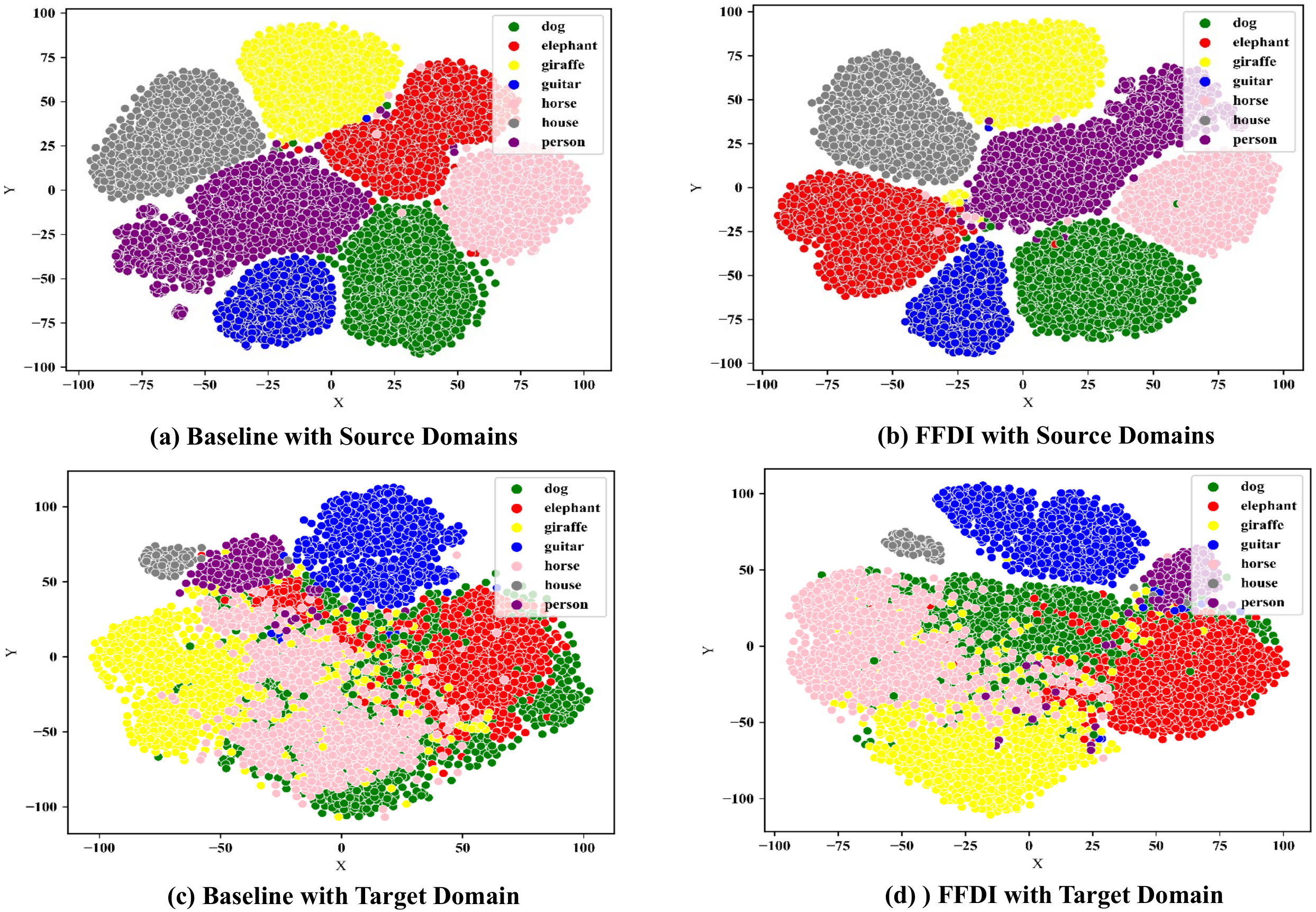}
    \caption{Results of feature visualization on PACS by t-SNE. We feed the output of the penultimate feature layer in the corresponding model that uses the sketch as the unknown domain to t-SNE. 
    }
    \label{fig:4}
\end{figure*}

\noindent\textbf{Impact of Different Components of FDAG:} Next, we will briefly analyze the proposed frequency-domain-based data augmentation method in this paper. Specifically, we use multiplicative noise and additive noise to perturb the phase and amplitude of the image, where multiplicative noise is the uniformly distributed random number with values between [0.5, 1.5] and additive noise is the Gaussian noise with the signal-to-noise ratio of 30 dB. The experimental results are shown in Tab.~$\ref{tab:6}$. We tried three types of data 
augmentation separately: ``none'' for no data augmentation, ``phase'' for perturbing the phase of the image, ``amplitude'' for perturbing the amplitude of the image, and ``amplitude + phase'' for perturbing the phase and amplitude of the image. We find that although the phase of the image preserves the semantic information of the object~\cite{xu2021}, appropriate perturbation of the phase can also improve the generalization of the network ($2\%$ and $1.5\%$ higher than the corresponding baseline on DeepAll and FFDI, respectively). The generalization performance of the network is further improved when we perturb both the amplitude and phase.

\noindent\textbf{Impact of Different Data Augmentation Methods:} We compare some common data augmentation with FADG, where ``standard'' indicates the common data augmentation methods in deep learning, including horizontal flipping, color jittering, random graying, random rotation, and a series of operations. ``t.-d. noise'' indicates direct perturbation on the time domain of the image (noise consistent with FDAG). ``AM'' indicates the method of ~\cite{xu2021} that exchanges the amplitude of different images. The results are as shown in Tab.~$\ref{tab:7}$. We find that frequency domain perturbation is more advantageous to enhancing the generalization of the network, which may be since FDAG can reduce the sensitivity of the model to frequency fluctuation, thus improving the robustness of the network in extracting high- and low-frequency features of the image. 

\noindent\textbf{Visual Analysis of Extracted Features:} To more vividly verify the robustness of our extracted features, we visualize the penultimate feature of baseline and FFDI using t-SNE. The visualization results are shown in Fig.~\ref{fig:4}. We can see that the features extracted by baseline in the source domain can be well separated, however, in the target domain, our proposed framework FFDI can extract more discriminative embedded features compared to baseline, which intuitively expresses the validity of the idea of interaction based on high and low-frequency features of the image.

\begin{figure}[t]
    \centering
    \includegraphics[width=6cm]{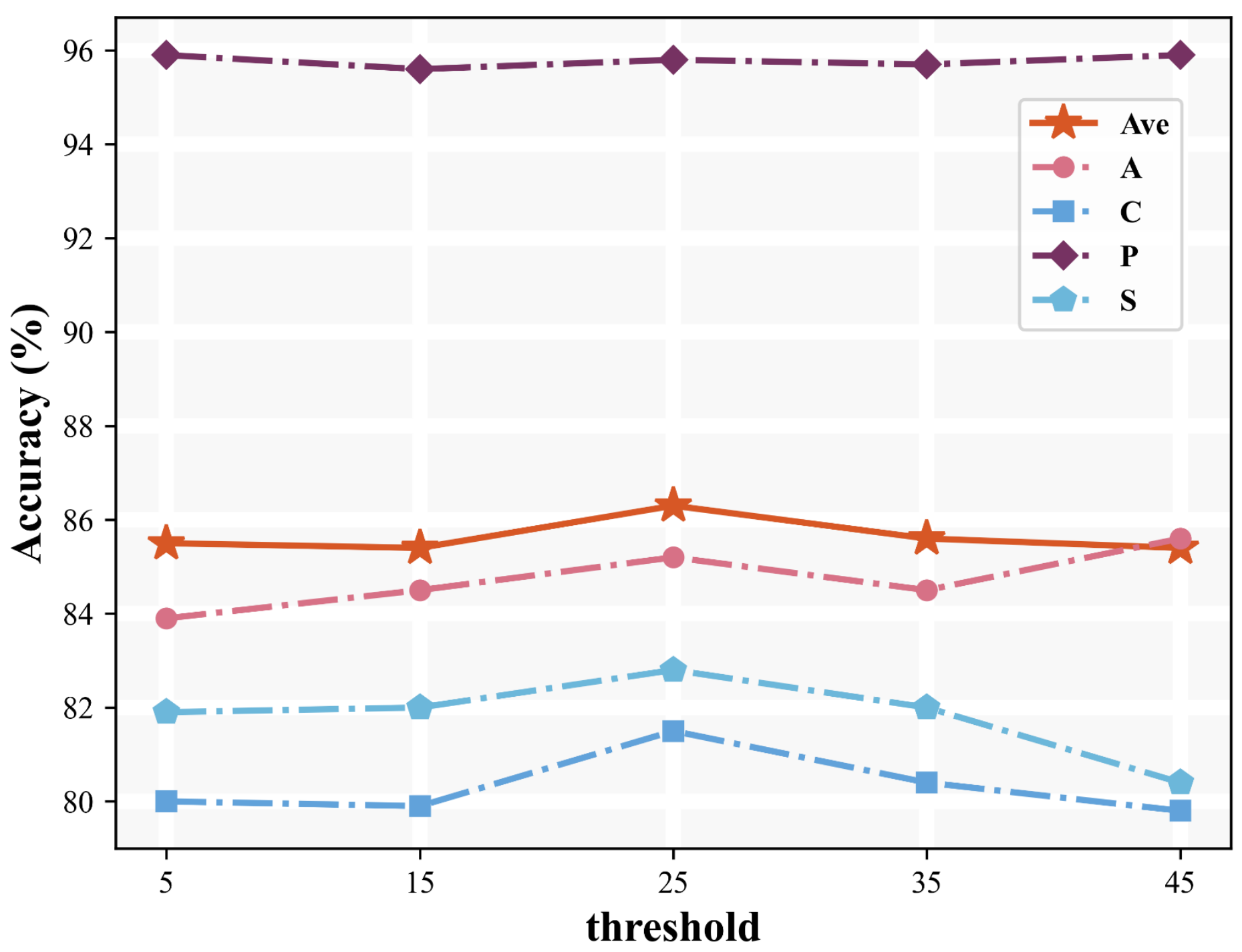}
    \caption{Experimental results for different frequency thresholds \textbf{r} on PACS with ResNet18. PACS stands for Photo, Art, Cartoon, and Sketch, respectively. The X- and Y-axis in the figure indicate the threshold and accuracy, respectively.}
    \label{fig:5}
\end{figure}

\noindent\textbf{Impact of Different Frequency Threshold:} We report the results for art, cartoon, photo, sketch, and average performance at different frequency threshold r in Fig.~\ref{fig:5}, respectively. The results show that FFDI achieves the best average performance on PACS dataset at r = 25. In addition, we find that the sketch domain is more sensitive to the value of r. As the percentage of high-frequency components decreases (as r increases), the test results in the sketch domain show a decreasing trend. The reasons for this phenomenon are due to the sketch image consisting of lines that are well described by high-frequency components. And the remaining three domains are relatively less affected by the threshold r and competitive results were obtained for each 
domains at r = 25.

\section{Conclusion}

In this study, we approach the domain generalization problem from a novel perspective of the image frequency domain, which uses low-frequency information to assist high-frequency information for identification. Specifically, we propose a new method named FFDI that disentangles the high- and low-frequency features of images and then fuses their helpful knowledge together with a simple yet effective information interaction mechanism (IIM). Furthermore, to achieve robust feature disentangling, we introduce a frequency-domain-based data augmentation technique to apply amplitude- and phase-wise disturbance to enrich the diversity of sample distribution. Finally, we evaluated the proposed method on three benchmark datasets with extensive experiments and reported state-of-the-art performance.

\begin{acks}
This work was supported in part by National Natural Science Foundation of
China (NSFC) No. 61922015, 62106022, U19B2036, and in part by Beijing Natural Science
Foundation Project No. Z200002.
\end{acks}

\clearpage

\bibliographystyle{ACM-Reference-Format}
\bibliography{sample-base}

\end{document}